\documentclass{article}
\usepackage[preprint]{neurips_2026}

\usepackage[utf8]{inputenc}
\usepackage[T1]{fontenc}
\usepackage{hyperref}
\usepackage{url}
\usepackage{booktabs}
\usepackage{amsfonts}
\usepackage{amsmath}
\usepackage{nicefrac}
\usepackage{microtype}
\usepackage{graphicx}
\usepackage{subcaption}
\usepackage{multirow}
\usepackage{pifont}
\usepackage[table]{xcolor}
\usepackage{caption}

\newcommand{\cmark}{\ding{51}}
\newcommand{\xmark}{\ding{55}}

\newcommand{\Description}[1]{}

\title{SentiAvatar: Towards Expressive and \\ Interactive Digital Humans}

\author{
  Chuhao Jin$^{1,2,\ast}$ \quad
  Rui Zhang$^{2,\ast}$ \quad
  Qingzhe Gao$^{2}$ \quad
  Haoyu Shi$^{3}$ \quad
  Dayu Wu$^{2}$ \\[3pt]
  \textbf{Yichen Jiang}$^{2}$ \quad
  \textbf{Yihan Wu}$^{1}$ \quad
  \textbf{Ruihua Song}$^{1,\dagger}$ \\[6pt]
  $^{1}$Gaoling School of Artificial Intelligence, Renmin University of China \\
  $^{2}$SentiPulse \\
  $^{3}$College of Computer Science, Inner Mongolia University \\[4pt]
  {\small $^\ast$Equal contribution.\ Chuhao Jin led this project.\quad $^\dagger$Corresponding author.} \\[2pt]
  \url{https://sentiavatar.github.io}
}

\begin{document}

\maketitle

\begin{abstract}
We present \textbf{SentiAvatar}, a framework for building expressive interactive 3D digital humans, and use it to create \textbf{SuSu}, a virtual character that speaks, gestures, and emotes in real time.
Achieving such a system remains challenging, as it requires jointly addressing three key problems: the lack of large-scale high-quality multimodal data, robust semantic-to-motion mapping, and fine-grained frame-level motion-prosody synchronization. 
To solve these problems, first, we build \textbf{SuSuInterActs} (21K clips, 37 hours), a dialogue corpus captured via optical motion capture around a single character with synchronized speech, full-body motion, and facial expressions. 
Second, we pre-train a \textbf{Motion Foundation Model} on 200K+ motion sequences, equipping it with rich action priors that go well beyond the conversation. 
We then propose an audio-aware \textbf{plan-then-infill} architecture that decouples sentence-level semantic planning from frame-level prosody-driven interpolation, so that generated motions are both semantically appropriate and rhythmically aligned with speech. 
Experiments show that SentiAvatar achieves state-of-the-art on both SuSuInterActs (R@1 43.64\%, nearly 2$\times$ the best baseline) and BEATv2 (FGD 4.941, BC 8.078), producing 6\,s of output in 0.3\,s with unlimited multi-turn streaming. The source code, model, and dataset are available at \url{https://sentiavatar.github.io}.

\end{abstract}

\begin{figure}[t]
    \centering
    \includegraphics[width=\textwidth]{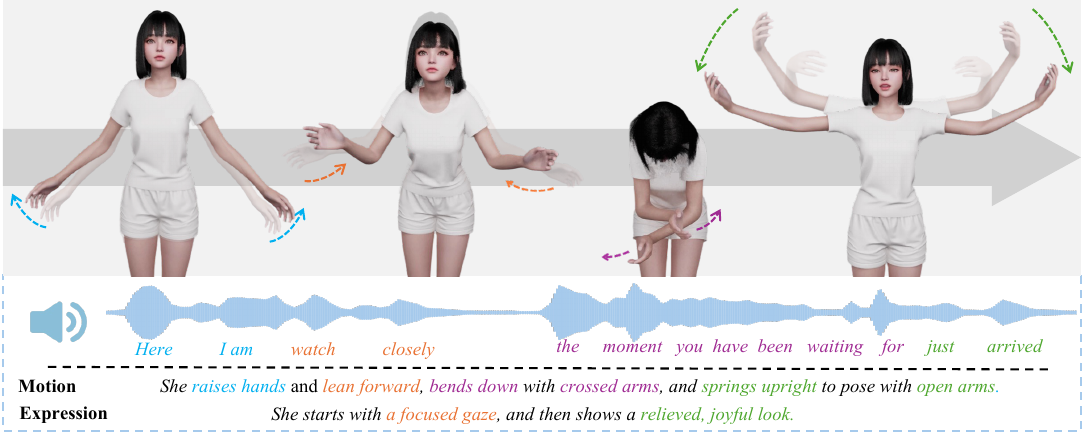}
    \caption{SentiAvatar generates high-quality 3D human motion and expression, which are semantically aligned and frame-level synchronized. The same color indicates the same time step.}
    \label{fig:teaser}\vspace{-10pt}
\end{figure}

\section{Introduction}
\label{sec:introduction}

People communicate with more than words.
A shrug carries helplessness; a nod signals agreement; a raised eyebrow conveys doubt.
These non-verbal behaviors---gestures, postures, facial expressions---are integral to interaction, not mere accessories to speech.
Giving 3D digital characters this expressiveness matters across domains: virtual assistants need it to build trust, robots need it to cooperate with humans, and games need it to make characters more vivid.
Current research focuses on generating motions from text prompts~\cite{guo2020action2motion,petrovich2021action,zhang2023t2m,jiang2023motiongpt,tevet2023human,guo2024momask} or producing gestures from speech audio~\cite{liu2024emage,yoon2020speech,liu2022learning,zhu2023taming,ao2022rhythmic}.
Few works study how to generate motions that are semantically consistent with dialogue context and intended utterances in interactive scenarios.

Building an interactive digital human that naturally gestures and emotes during real dialogue poses three challenges:
(1)~\emph{Lack of high-quality data}---full-body motion and facial expressions during conversational interaction. Most text-to-motion datasets~\cite{Guo_2022_CVPR,BABEL:CVPR:2021,lin2023motionx} only pair action descriptions with motion sequences, without dialogue or interaction context; speech-driven datasets such as BEAT~\cite{liu2022beat} and BEATv2~\cite{liu2024emage} are built from monologue-style presentations rather than bidirectional conversation.
(2)~\emph{Insufficient semantic-to-motion alignment}---when motion descriptions are complex or lengthy, generated motions often miss certain actions.  (bridging the gap between discrete high-level semantic intents and continuous motion dynamics is difficult. When descriptions are complex, existing models often struggle to faithfully execute all intended actions)
(3)~\emph{Under-explored prosody alignment in dialogue}---existing co-speech methods are trained on single-speaker presentation data, which differs from the motion patterns that arise in person-to-person dialogue.

We present \textbf{SentiAvatar}, as shown in Figure~\ref{fig:teaser}, a framework for building expressive interactive 3D digital humans, and use it to create \textbf{SuSu}, a virtual character that speaks, gestures, and emotes in real time.
We tackle all three challenges in a unified pipeline.
For \emph{data}, we design multi-scenario dialogue scripts around SuSu with behavior annotations and have professional actors perform them in an optical motion-capture studio, capturing synchronized speech, full-body motion, and facial expressions to form \textbf{SuSuInterActs} (21K clips, 37 hours).
To improve motion generalization and diversity, we pre-train a \textbf{Motion Foundation Model} on 200K+ sequences aggregated from public corpora and the Tencent Hunyuan Motion Model, giving SuSu rich action priors that go well beyond the conversational domain.
With data and motion knowledge in place, we propose a new framework, \textbf{SentiAvatar}, that achieves both semantic alignment between text prompts and generated motions and rhythmic alignment between generated motions and speech.
Our audio-aware \textbf{plan-then-infill} architecture decouples these two objectives: the fine-tuned foundation model takes behavior labels and audio tokens as input and generates semantically aligned sparse keyframes; an Infill Transformer then fills in the intermediate frames conditioned on boundary keyframes and frame-level speech features, producing motions that are both semantically appropriate and rhythmically aligned.
The resulting system achieves state-of-the-art on both SuSuInterActs and BEATv2~\cite{liu2024emage}, producing 6\,s of output in 0.3\,s with unlimited multi-turn streaming.

Our contributions are three-fold:
\textbf{First}, we open-source an interactive 3D digital human framework: SuSu speaks, gestures, and emotes in real-time dialogue.
\textbf{Second}, we construct SuSuInterActs, a 21K-clip, 37-hour multimodal dialogue corpus with synchronized speech, behavior-annotated text, full-body motion, and facial expressions.
\textbf{Third}, we propose the plan-then-infill architecture and a Motion Foundation Model pre-trained on 200K+ sequences, achieving state-of-the-art on both SuSuInterActs (R@1 43.64\%, nearly 2$\times$ the best baseline) and BEATv2 (FGD 5.301→4.941, BC 7.971→8.078, improving the best prior results on both metrics).
\begin{table*}[t]
\centering
\scalebox{0.83}{
\begin{tabular}{l c c c c c c c c}
\toprule
Dataset 
& \#Samples 
& \shortstack[c]{Multi-\\turn} 
& \shortstack[c]{Role-playing\\Dialogue} 
& \shortstack[c]{Action\\Description} 
& Audio 
& Face 
& \shortstack[c]{Body\\Motion} 
& \shortstack[c]{Hands\\Motion} \\
\midrule
HumanML3D~\cite{Guo_2022_CVPR} 
& 16k 
& \xmark 
& \xmark 
& \cmark 
& \xmark 
& \xmark 
& \cmark 
& \xmark \\

Motion-X~\cite{lin2023motionx} 
& 95k 
& \xmark 
& \xmark 
& \cmark 
& \xmark 
& \xmark 
& \cmark 
& \cmark \\

BEAT~\cite{liu2022beat} 
& 2.5k 
& \xmark 
& \xmark 
& \cmark 
& \cmark 
& \cmark 
& \cmark 
& \xmark \\

SnapMoGen~\cite{snapmogen2025} 
& 20k 
& \cmark 
& \xmark 
& \cmark 
& \xmark 
& \xmark 
& \cmark 
& \xmark \\

InterX~\cite{xu2023interx} 
& 31k 
& \xmark 
& \xmark 
& \cmark 
& \xmark 
& \xmark 
& \cmark 
& \cmark \\

InterHuman~\cite{liang2024intergen} 
& 6k 
& \xmark 
& \xmark 
& \cmark 
& \xmark 
& \xmark 
& \cmark 
& \xmark \\

CharacterEval~\cite{tu2024charactereval} 
& 11k 
& \cmark 
& \cmark 
& \cmark 
& \xmark 
& \xmark 
& \xmark 
& \xmark \\

\midrule
\textbf{SuSuInterActs (Ours)} 
& \textbf{21k} 
& \textbf{\cmark} 
& \textbf{\cmark} 
& \textbf{\cmark} 
& \textbf{\cmark} 
& \textbf{\cmark} 
& \textbf{\cmark} 
& \textbf{\cmark} \\
\bottomrule
\end{tabular}}
\caption{Comparison between SuSuInterActs and existing motion and role-playing dialogue datasets. SuSuInterActs uniquely combines multi-turn dialogue, role-conditioned interactions, and synchronized multimodal signals including text, audio, facial expressions, and full-body motion with hands.}
\label{tab:dataset_comparison}\vspace{-10pt}
\end{table*}

\section{Related Work}
\label{sec:related}

\subsection{Text-driven Human Motion Generation}
\label{sec:related_t2m}

Text-driven motion generation synthesizes body movements from natural language descriptions. Early VAE-based methods~\cite{guo2020action2motion,petrovich2021action,tevet2022motionclip} generate diverse action-conditioned motions, with HumanML3D~\cite{Guo_2022_CVPR} providing the standard benchmark. Diffusion-based approaches, such as MDM~\cite{tevet2023human}, FineMoGen~\cite{zhang2023finemogen}, SALAD~\cite{hong2025salad}, EnergyMoGen~\cite{zhang2025energymogen}, MoLA~\cite{uchida2024mola}, MotionFlow~\cite{11329478}---improve quality, controllability. DartControl~\cite{Zhao:DartControl:2025} and ActionPlan~\cite{nazarenus2026actionplan} support real-time streaming. Discrete-token methods~\cite{zhang2023generating,guo2024momask,pinyoanuntapong2024mmm,pinyoanuntapong2025bamm,yuan2024mogents,light-t2m,wang2026temporal,dang2026segmo} achieve competitive results via VQ-VAE and masked modeling. LLMs have also been adopted as motion generators: Motion-Agent~\cite{Wu2024MotionAgentAC}, MotionGPT3~\cite{zhu2025motiongpt3humanmotionsecond}, MG-MotionLLM~\cite{wu2025mg}, Motion-R1~\cite{ouyang2025motion}, SMooGPT~\cite{zhong2025smoogpt}, and PlanMoGPT~\cite{jin2025planmogpt}. These methods all generate motion from static text and do not model speech. We extend the discrete-token LLM paradigm to dialogue-driven generation, where the LLM plans sparse keyframes that a prosody-aware Infill Transformer densifies.

\subsection{Speech-driven Gesture and Expression Generation}
\label{sec:related_gesture}

Co-speech gesture generation produces body movements synchronized with speech. Early data-driven approaches~\cite{ginosar2019gestures,ahuja-etal-2020-gestures} learn from in-the-wild video. Subsequent work improves quality through hierarchical modeling~\cite{liu2022learning}, rhythm-aware segmentation~\cite{ao2022rhythmic}, diffusion-based synthesis~\cite{zhu2023taming,alexanderson2023listen}, VQ-VAE with phase guidance~\cite{yang2023qpgesture}, CLIP-based style control~\cite{ao2023gesturediffuclip}, and contrastive speech-motion pre-training~\cite{deichler2023diffusionbased,liu2022audio}. Holistic methods jointly cover body, hands, and face: TalkShow~\cite{yi2022generating} and EMAGE~\cite{liu2024emage} use compositional VQ-VAEs, DiffSHEG~\cite{chen2024diffsheg} couples expression and gesture, and M3G~\cite{yin2025m3g} captures multi-granular gesture patterns. Interactive and conversational settings are addressed by CoDiffuseGesture~\cite{10448208}, DiffuGesture~\cite{zhao2023diffugesture}, Co3Gesture~\cite{qi2025comathbfgesture}, and MIBURI~\cite{mughal2026miburi}. A few works inject text via LLM-based retrieval~\cite{zhang2024semantic}, multimodal language models~\cite{chen2024language,chen2025meco}, or text guidance~\cite{peng2024t3m}. Despite this progress, most methods treat gestures as a low-level reflex of speech rhythm without sentence-level semantic planning. SentiAvatar decouples semantic planning from prosodic infilling, jointly generates body motion and facial expressions, and leverages a 200K-sequence motion foundation model.

\subsection{Motion Representation, Foundation Models, and Datasets}
\label{sec:related_repr}

VQ-VAE~\cite{zhang2023generating,yang2023qpgesture, guo2024momask} provides discrete motion representations; we adopt R-VQVAE~\cite{guo2024momask} to connect the LLM planner and Infill Transformer. Scaling motion models via large-scale pre-training is an emerging direction: Kimodo~\cite{Kimodo2026} trains on 700 hours of mocap, GENMO~\cite{genmo2025} unifies estimation and generation, ViMoGen~\cite{lin2025quest} transfers video priors with a 228K-sample dataset, and VimoRAG~\cite{xu2025vimorag} augments motion LLMs with video retrieval---motivating our pre-training on 200K+ heterogeneous sequences. For datasets, HumanML3D~\cite{Guo_2022_CVPR}, BABEL~\cite{BABEL:CVPR:2021}, Motion-X~\cite{lin2023motionx}, and KinMo~\cite{kinmo2025kinematicawarehumanmotion} provide text-motion benchmarks; BEAT/BEAT2~\cite{liu2022beat,liu2024emage} and MM-Conv~\cite{deichler2024mm} cover co-speech data; AIST++~\cite{li2021learn} and SoulDance~\cite{li2025souldance} target dance. Recent large-scale efforts include Embody 3D~\cite{embody3d} (500h), Nymeria~\cite{ma24eccv} (300h), SnapMoGen~\cite{snapmogen2025} (20K clips), Seamless Interaction~\cite{agrawal2025seamless} (4,000h), InterHuman~\cite{liang2024intergen}, CHI3D~\cite{fieraru2020three}, Intend to Move~\cite{umagami2025intend}, and PersonaBooth~\cite{kim2025personabooth}. Existing co-speech datasets are mostly based on non-conversational TED Talks, or lack synchronized facial expressions and semantic behavior annotations. Our SuSuInterActs fills this gap: 21K clips, 37 hours of Chinese dialogue with synchronized speech, full-body motion, facial expressions, and per-utterance behavior labels for single-character persona modeling.

\begin{figure*}
    \centering
    \includegraphics[width=1.0\linewidth]{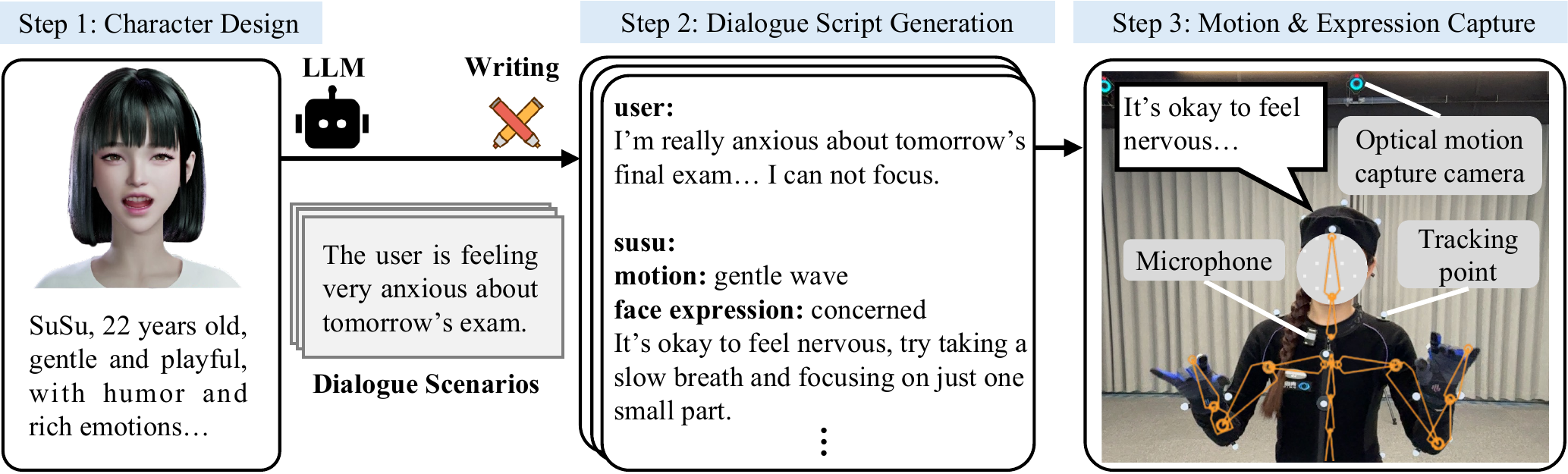}
    \caption{Overview of SuSuInterActs data pipeline. (1) Character design for a consistent persona. (2) LLM script generation with explicit motion/expression tags. (3) Multimodal capture of synchronized speech, full-body motion, and facial expressions.}
    \label{fig:data_collection}\vspace{-10pt}
\end{figure*}

\section{Constructing SuSuInterActs Dataset}
\label{sec:dataset}

We construct \textbf{SuSuInterActs}, a multimodal Chinese conversation dataset centered on a single virtual character.
Each sample pairs temporally aligned speech audio, dialogue text, full-body motion, and facial expressions.
Formally:
\begin{equation}
\mathcal{D} = \{ (\mathcal{C}_i, \mathcal{S}_i) \}_{i=1}^{N},
\end{equation}
where $\mathcal{C}_i$ is an interaction scenario and $\mathcal{S}_i$ the corresponding dialogue session.
Since the dataset focuses on a single character, the role definition is shared across all samples.

As shown in Figure~\ref{fig:data_collection}, the collection proceeds in three stages: character design, script generation, and performance capture.

\noindent\textbf{Character Design.}
SuSu is a 22-year-old virtual character with a warm yet playful personality.
Her emotional range---curiosity, encouragement, teasing, bashfulness---ensures diverse motion and expression patterns in the captured data.
The full character profile is provided in Appendix~\ref{app:character}.

\noindent\textbf{Dialogue Script Generation.}
We design dialogue scenarios spanning everyday topics---casual chat, storytelling, emotional support, playful banter---to elicit diverse emotional states and interaction dynamics.
We then prompt a large language model to generate a multi-turn dialogue script based on the given scenario $\mathcal{C}_i$.
Each utterance turn is annotated as:
\begin{equation}
\tau_{i,t} = \big( \mathbf{e}_{i,t},\ \mathbf{a}_{i,t},\ \mathbf{u}_{i,t} \big),
\end{equation}
where $\mathbf{e}_{i,t}$ is a facial expression label (e.g., ``raised eyebrows''), $\mathbf{a}_{i,t}$ a body action label (e.g., ``tilts head''), and $\mathbf{u}_{i,t}$ the spoken text.
All labels use concrete, performable phrases formatted as inline tags.
Each script is manually reviewed to filter impractical actions.

\noindent\textbf{Motion \& Expression Capture.}
Professional actors study the character profile and script before each session; light improvisation consistent with SuSu's personality is permitted.
Full-body motion including hand gestures is recorded via optical motion capture.
Facial expressions are captured with iPhone ARKit as 51-dim blendshape coefficients.
Speech audio is also recorded.
Each turn yields:
\begin{equation}
\mathbf{f}_{i,t} = \big( \mathbf{m}_{i,t},\ \mathbf{b}_{i,t} \big),
\end{equation}
where $\mathbf{m}_{i,t} \in \mathbb{R}^{L \times 393}$ is the full-body pose sequence (6D rotation for 63 joints: 25 body + 19$\times$2 hand) and $\mathbf{b}_{i,t} \in \mathbb{R}^{L \times 51}$ is the blendshape sequence.
Raw motion is retargeted onto SuSu's skeleton, and all modalities are temporally aligned to 20\,FPS after quality filtering.

\begin{table}
    \centering
    \caption{Statistics of the SuSuInterActs dataset.}
    \scalebox{0.85}{
    \begin{tabular}{l c}
    \toprule
    Statistic & Value \\
    \midrule
    \# Total Samples & 21,133 \\
    Total Duration & 36.9 hours \\
    Avg. Duration per Sample & 6.3 s \\
    Avg. Characters per Utterance & 18.7 \\
    \# Samples w/ Non-default Expression & 9,412 \\
    \# Samples w/ Non-default Action & 14,278 \\
    \# Samples w/ Any Non-default Label & 18,696 \\
    Total Motion Frames & 2,656,484 \\
    Avg. Frames per Sample & 125.7 \\
    \# Samples w/ Facial Data & 12,367 \\
    \midrule
    Frame Rate & 20 FPS \\
    Motion Representation & 6D rotation, 63 joints \\
    Face Representation & 51-dim ARKit blendshape \\
    \bottomrule
    \end{tabular}}\vspace{-10pt}
    \label{tab:susu_stats}
\end{table}

\paragraph{Dataset Analysis.}
Table~\ref{tab:susu_stats} summarizes the dataset statistics. SuSuInterActs contains 21K samples totaling 37 hours, with an average of 6.3\,s and 18.7 Chinese characters per turn.
Every sample carries expression and action labels; among them, 14K have non-default actions and 9K have non-default expressions.
Facial blendshape data is available for 12K samples.
SuSuInterActs is the first dataset that combines multi-turn dialogue, role-conditioned interactions, speech audio, full-body motion with hands, and facial expressions in a single corpus.

\paragraph{Dataset Comparison}
Table~\ref{tab:dataset_comparison} compares SuSuInterActs with representative motion and dialogue datasets.
Co-speech datasets such as BEAT~\cite{liu2022beat} provide audio-motion pairs but lack multi-turn interaction.
Text-to-motion datasets such as HumanML3D~\cite{Guo_2022_CVPR} and Motion-X~\cite{lin2023motionx} pair action descriptions with body motion, yet contain no audio or conversational grounding.
Interaction datasets such as InterX~\cite{xu2023interx} and InterHuman~\cite{liang2024intergen} capture multi-person motion without speech or role conditioning.
Role-playing dialogue datasets such as CharacterEval~\cite{tu2024charactereval} support multi-turn role-based dialogue but are text-only.
SuSuInterActs is the first to unify all these dimensions---multi-turn dialogue, role conditioning, speech audio, full-body motion with hands, and facial expressions---in a single corpus.
Focusing on one character further yields consistent behavioral patterns, benefiting character-specific generation.

\section{Method}
\label{sec:method}

\subsection{System Overview}
\label{sec:overview}

\begin{figure}[t]
    \centering
    \includegraphics[width=\linewidth]{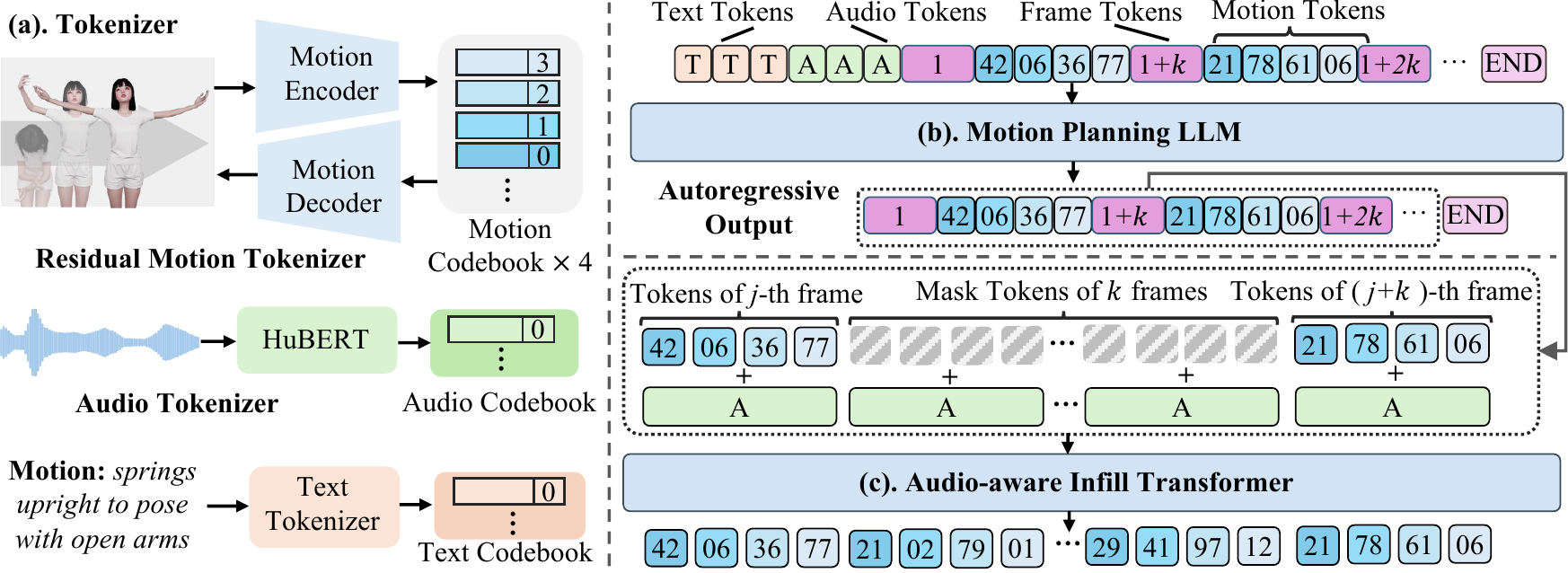}
    \caption{Overview of SentiAvatar. (a) Multi-modal inputs are quantized into tokens via encoders. The body pathway employs a hierarchical generation: (b) LLM planner predicts sparse keyframe tokens to capture high-level dialogue content, followed by (c) an audio-aware Infill Transformer for dense, prosody-driven interpolation to ensure fine-grained temporal synchronization.}
    \label{fig:system_overview}\vspace{-10pt}
\end{figure}

Dialogue-driven motion generation requires alignment at two temporal scales: sentence-level \emph{semantic alignment} maps dialogue intent to action types (e.g., ``shrug,'' ``nod''), while frame-level \emph{prosody alignment} synchronizes motion dynamics with speech rhythm.
SentiAvatar decomposes generation into two stages (Figure~\ref{fig:system_overview}).
In Stage~I, an LLM-based \emph{Motion planner} takes behavior labels and sparse audio tokens as input and produces keyframe motion tokens at interval $t$.
In Stage~II, a lightweight \emph{Audio-aware Infill Transformer} fills the $t{-}1$ intermediate frames conditioned on dense audio features.
R-VQVAE discrete tokens serve as the unified interface between the two stages.
For facial expressions, a separate Face Infill Transformer generates face tokens directly from audio.
Since HuBERT features already encode rich linguistic and prosodic semantics, and facial movements are tightly coupled with speech, no additional LLM planning is needed for this pathway.

The following subsections describe discrete representation (\S\ref{sec:representation}), semantic planning (\S\ref{sec:llm_planning}), motion infilling (\S\ref{sec:infill}), face generation (\S\ref{sec:face}), motion foundation model (\S\ref{sec:foundation}), and inference pipeline (\S\ref{sec:inference}).

\subsection{Discrete Representation}
\label{sec:representation}

\noindent\textbf{Residual Motion Tokenizer.}
We adopt a Residual VQ-VAE (R-VQVAE)~\cite{guo2024momask} to convert continuous joint rotations into discrete tokens.
A 1D convolutional encoder downsamples temporally by 2$\times$; 4-layer residual quantization with codebook size 512 per layer produces token groups $(r^1, r^2, r^3, r^4)$ per time step.
Layer-specific offsets yield a unified vocabulary of $512 \times 4 = 2{,}048$ IDs:
\begin{equation}
\hat{r}^k = r^k + 512 \times (k-1), \quad k \in \{1,2,3,4\}.
\end{equation}
The decoder reconstructs continuous motion from predicted tokens.
Following previous works~\cite{guo2024momask}, the R-VQVAE is trained by combines reconstruction, velocity, commitment, and position losses.

\noindent\textbf{Audio Representation.}
We extract features from a HuBERT model at 50\,FPS, downsampled to 20\,FPS.
Two forms are used:
(1)~\emph{discrete tokens} via K-means clustering, subsampled at the keyframe interval $t$ for the LLM;
(2)~\emph{continuous} 768-dim feature vectors at 20\,FPS for the Infill Transformers.

\subsection{LLM-based Semantic Planning}
\label{sec:llm_planning}

Generating motion from dialogue requires understanding high-level intent: the model must know \emph{what} action to perform before deciding \emph{how} to perform it frame by frame.
We cast this as a planning problem: an LLM reads the motion label and coarse audio context, then outputs a sparse sequence of keyframe motion tokens that sketch the intended action trajectory.
By operating at the keyframe level, the LLM focuses on semantic correctness---selecting the right action types and their temporal layout---without being burdened by frame-level dynamics.
This sparse plan also provides strong boundary constraints for the subsequent infilling stage.

\noindent\textbf{Task Formulation.}
Given a motion label $\mathbf{T}$ (e.g., \texttt{shrug helplessly}) and sparse audio tokens $\{a_1, a_{1+t}, a_{1+2t}, \ldots\}$ sampled at interval $t$, the LLM generates sparse keyframe motion tokens:
\begin{align}
\text{Input:} &\quad \mathbf{T} \oplus [a_1][a_{1+t}][a_{1+2t}] \cdots, \\
\text{Output:} &\quad [S_t][\mathbf{r}_1][\mathbf{r}_{1+t}][\mathbf{r}_{1+2t}] \cdots,
\end{align}
where $[S_t]$ is the keyframe step, and each $\mathbf{r}_i = (r^1_i, r^2_i, r^3_i, r^4_i)$ is a 4-token residual group.
The sparse audio tokens provide coarse temporal context---overall utterance rhythm, pause locations, emphasis patterns---that helps the LLM place keyframes at appropriate moments even before dense prosody alignment.

\noindent\textbf{Training.}
The model is initialized from the Motion Foundation Model (\S\ref{sec:foundation}) and fine-tuned on SuSuInterActs via full-parameter SFT.
Pre-trained motion priors provide a strong initialization.
A \emph{continuation mode} supports seamless multi-turn generation by prepending the last two keyframe pairs from the previous utterance as context (details in Appendix~\ref{app:continuation}).

\subsection{Audio-conditioned Motion Infilling}
\label{sec:infill}

The LLM-based planner produces keyframes that capture \emph{what} actions to perform, but the gaps between keyframes lack fine-grained dynamics---the subtle accelerations, hesitations, and rhythmic gestures that make motion look natural and speech-synchronized.
The Audio-aware Infill Transformer addresses this by conditioning on dense, frame-level audio features to fill intermediate frames, injecting prosody-driven nuance into the coarse plan.

For a sliding window of $t{+}1$ frames, the two boundary keyframes are known and the $t{-}1$ interior frames are masked:
\begin{equation}
\text{Input:} \quad [\mathbf{r}_i]\underbrace{[\texttt{M}]\cdots[\texttt{M}]}_{t-1}[\mathbf{r}_{i+t}] + \mathbf{h}_{i:i+t},
\end{equation}
where $\mathbf{h} \in \mathbb{R}^{(t+1) \times 768}$ are frame-level HuBERT features.
The boundary keyframes anchor the motion trajectory while the audio features guide the dynamics---stressed syllables can trigger sharper movements, pauses can induce holds.
The model is a Transformer Encoder (8 layers, 16 heads, dim 512, $\sim$38.5M parameters).
Audio features are projected via a two-layer MLP and fused with token embeddings by element-wise addition.
Learnable positional encodings distinguish temporal positions within the window.

\noindent\textbf{Training.}
Within the $t{-}1$ interior frames, a random subset of tokens is masked per sample; unmasked tokens have a 10\% probability of being replaced with random values to improve robustness.
Boundary frames are always known and excluded from the loss.

\subsection{Facial Expression Generation}
\label{sec:face}

As noted in \S\ref{sec:overview}, HuBERT features encode rich linguistic and prosodic information---phoneme identity, stress, and intonation---which directly drives facial movements: lip shapes follow phonemes, eyebrow raises align with emphasis, smiles correlate with vocal warmth.
Sentence-level planning is therefore unnecessary.

We use a Face R-VQVAE (2-layer, codebook size 512, vocabulary of $1{,}024$ IDs) to tokenize 51-dim ARKit sequences, producing 2 tokens per frame.
A Face Infill Transformer, sharing the same architecture and training procedure as the body counterpart (\S\ref{sec:infill}), generates face tokens directly from audio features.
The face pathway runs in parallel with the body pathway at inference time.

\subsection{Motion Foundation Model}
\label{sec:foundation}

To broaden the motion prior beyond conversational data, we pre-train a Motion Foundation Model on 200K+ sequences (676 hours) from diverse sources (Table~\ref{tab:pretrain_data}): EmbodyAI (conversations, daily activities), SnapMoGen (general motion), Motion-X (daily, dance, sports), and 15K sequences distilled from the Tencent Hunyuan Motion Model via systematic prompt engineering (details in Appendix~\ref{app:hunyuan}).
All data is retargeted to a unified skeleton and tokenized by the Motion R-VQVAE.

\begin{table}[t]
    \centering
    \caption{Pre-training data for the Motion Foundation Model. All sequences are retargeted to a unified skeleton and tokenized by the Motion R-VQVAE.}
    \scalebox{0.95}{
    \begin{tabular}{l r r c}
    \toprule
    \textbf{Source} & \textbf{\# Seqs} & \textbf{Hours} & \textbf{Type} \\
    \midrule
    EmbodyAI & 84,010 & 467.7 & Conv., Daily \\
    SnapMoGen & 20,450 & 43.7 & General \\
    Motion-X & 81,084 & 144.2 & Daily, Dance, Sports \\
    Hunyuan Distill. & 15,000 & 20.8 & Atomic \& Composite \\
    \midrule
    \textbf{Total} & \textbf{200,544} & \textbf{676.4} & --- \\
    \bottomrule
    \end{tabular}}
    \label{tab:pretrain_data}\vspace{-10pt}
\end{table}

The model is initialized from Qwen-0.5B with motion and audio tokens added to the vocabulary.
Pre-training uses a text-to-motion autoregressive objective:
\begin{equation}
P(\mathbf{r}_{1:N} \mid \mathbf{T}) = \prod_{i=1}^{N} P(r^1_i, r^2_i, r^3_i, r^4_i \mid \mathbf{T}, \mathbf{r}_{1:i-1}),
\end{equation}
where $\mathbf{T}$ is a Chinese motion description and $N$ the number of token groups.
This gives the downstream planner a broad motion vocabulary and faster convergence on SuSuInterActs.

\subsection{Inference Pipeline}
\label{sec:inference}

At inference, HuBERT extracts audio features from input speech; K-means quantization produces discrete tokens $a_i$ for the LLM, while continuous features $\mathbf{h}_i$ feed the Infill Transformers.
The \textbf{body pathway} runs the full two-stage pipeline: the LLM generates sparse keyframe tokens from the motion label $\mathbf{T}$ and audio tokens; the body Infill Transformer then interpolates dense frames via iterative refinement---at each of 6 steps, the most confident predictions are accepted and remaining masked positions continue to the next step; finally, the Motion R-VQVAE decodes tokens into continuous joint rotations.
The \textbf{face pathway} bypasses the LLM: the Face Infill Transformer generates face tokens from audio features, decoded by the Face R-VQVAE into blendshape coefficients.
Both pathways share HuBERT features and run in parallel.
For multi-turn dialogue, the LLM's continuation mode and the Infill Transformer's sliding window handle cross-utterance transitions.
End-to-end latency is $\sim$0.3\,s for $\sim$6\,s of output, supporting real-time streaming.

\begin{table*}[t]
    \centering
    \caption{Quantitative comparison on SuSuInterActs. \textbf{Bold}: best;  $\uparrow$/$\downarrow$: higher/lower is better. ESD in seconds. ``$^\dagger$'' indicates T2M-GPT variants with token-by-token autoregressive generation.}
    \label{tab:susu_main}
    \scalebox{0.92}{
    \begin{tabular}{l c c c c c c c}
    \toprule
    \textbf{Method} & \textbf{Condition} & \textbf{R@1}$\uparrow$ & \textbf{R@2}$\uparrow$ & \textbf{R@3}$\uparrow$ & \textbf{FID}$\downarrow$ & \textbf{ESD}$\downarrow$ & \textbf{Diversity}$\uparrow$ \\\midrule
    Real Motion  & ---  & 62.20 & 73.56 & 78.70 & 0.000 & 0.308 & 22.61 \\
    \midrule
    \multicolumn{8}{l}{\emph{Audio-only methods}} \\
    EMAGE~\cite{liu2024emage}    & Audio    & 5.00  & 9.40  & 13.32 & 441.6 & 0.606 & 12.92 \\
    A2M-GPT$^\dagger$  & Audio & 8.72 & 15.96 & 20.08   & 13.66 & 0.477   & 22.23 \\
    \midrule
    \multicolumn{8}{l}{\emph{Text-only methods}} \\
    HunYuan-Motion      & Text     & 5.21 & 8.59 & 11.9 & 352.56 & 0.708 & 16.92 \\
    T2M-GPT~\cite{zhang2023t2m}     & Text     & 23.12 & 30.49 & 35.43 & 67.78 & 0.721 & 20.65\\
    MoMask~\cite{guo2024momask}        & Text     & 34.55 & 46.58 & 54.29 & 36.25 & 0.471 & 22.03 \\
    \midrule
    \multicolumn{8}{l}{\emph{Audio + Text methods}} \\
    AT2M-GPT$^\dagger$         & Audio, Text & 27.52 & 36.11 & 41.38 & 18.491 & 0.503 & 22.36 \\\midrule
    \textbf{SentiAvatar (Ours)}         & Audio, Text & \textbf{43.64} & \textbf{54.94} & \textbf{61.84} & \textbf{8.912} & \textbf{0.456} & \textbf{22.41} \\
    \rowcolor{gray!10} \multicolumn{2}{l}{\textbf{Improvement(\%)}} & \textbf{+26.3} & \textbf{+17.9} & \textbf{+13.9} & \textbf{+34.8} & \textbf{+3.2} & \textbf{+0.2} \\
    \bottomrule
    \end{tabular}}\vspace{-5pt}
\end{table*}

\section{Experiments}
\label{sec:experiments}

\subsection{Experimental Setup}
\label{sec:exp_setup}

\paragraph{Datasets.}
We evaluate our method on two datasets:
(1)~\textbf{SuSuInterActs} (Section~\ref{sec:dataset}), our collected single-character Chinese dialogue motion dataset. We split it into 20,982 training, 710 validation, and 542 test samples.
(2)~\textbf{BEATv2}~\cite{liu2024emage}, a widely-used English co-speech gesture benchmark. We retrain our full pipeline on the BEATv2 and evaluate on its test split to verify cross-dataset generalization.

\paragraph{Evaluation Metrics.}
We adopt both objective and subjective metrics.
For objective evaluation on the SuSuInterActs dataset, following previous works~\cite{guo2024momask,zhang2023t2m}, we report \textbf{R@K} ($\uparrow$, $K \in \{1, 2, 3\}$), the text-to-motion retrieval recall measuring semantic alignment; \textbf{FID} ($\downarrow$), the Fr\'echet Inception Distance measuring overall motion quality; \textbf{ESD} ($\downarrow$), Event Sync Distance, a bidirectional event-level audio--motion synchronization metric that computes the average nearest temporal distance between detected audio onset events and motion velocity peak events---lower values indicate tighter synchronization (details in Appendix~\ref{app:esd}); and \textbf{Diversity} ($\uparrow$), the mean pairwise L2 distance among motion latent features.
On BEATv2, we follow the standard protocol and report \textbf{FGD} ($\downarrow$, Fr\'echet Gesture Distance), \textbf{BC} ($\uparrow$, Beat Consistency), and \textbf{Diversity}.
For subjective evaluation, we conduct a user study assessing semantic consistency, prosody synchronization, and overall quality (Section~\ref{sec:user_study}).

\paragraph{Baselines.}
We compare with methods spanning three conditioning paradigms.
For \emph{audio-only} conditioning, we include \textbf{EMAGE}~\cite{liu2024emage}, a state-of-the-art co-speech gesture generation method that uses masked audio transformers to synthesize full-body motion from speech audio.
For \emph{text-only} conditioning, we include \textbf{T2M-GPT}~\cite{zhang2023t2m}, a GPT-based text-to-motion method, and \textbf{MoMask}~\cite{guo2024momask}, a masked transformer-based text-to-motion method.
Since the SuSuInterActs dataset is in Chinese, the original English text encoders used by T2M-GPT and MoMask are not directly applicable. We therefore replace their text encoders with \texttt{bert-base-chinese} for fair comparison, ensuring that these baselines can properly encode our Chinese labels.
We also include \textbf{HunYuan-Motion}~\cite{hymotion2025}, a recent large-scale text-to-motion generation model, as an industry-level zero-shot baseline.
For \emph{audio + text} conditioning, we construct two T2M-GPT variants that extend the original text-only GPT architecture with additional input modalities: \textbf{A2M-GPT} replaces text with audio tokens as the conditioning signal, while \textbf{AT2M-GPT} uses both audio tokens and text labels. Both methods follow the same token-by-token autoregressive generation paradigm as T2M-GPT. For fair comparison, A2M-GPT and AT2M-GPT use the same Qwen-0.5B backbone as our method. These variants help isolate the contribution of our hierarchical two-stage design from the effect of conditioning modalities alone.
All baselines (except HunYuan-Motion) are retrained using their official codebases.

\paragraph{Implementation Details.}
SentiAvatar is trained in three stages on 8$\times$A100 GPUs:
(1)~\emph{R-VQVAE}: The motion and face R-VQVAEs are trained independently with batch size 128 for 100 epochs.
(2)~\emph{Motion Foundation Model}: The Qwen-0.5B backbone is pre-trained on the aggregated 200K motion sequences (Table~\ref{tab:pretrain_data}) for 10 epochs with per-GPU batch size 128.
(3)~\emph{SFT}: The pre-trained model is fine-tuned on the SuSu training set for 10 epochs with the same batch size configuration.
The Infill Transformers (body and face) are trained with a batch size 1024 for 100 epochs.
We use the AdamW optimizer with a learning rate of 1e-4 and a cosine annealing schedule.


\begin{table}[t]
    \centering
    \caption{Comparison on BEATv2. Baseline results from original publications.}
    \label{tab:beatv2_main}
    \scalebox{0.90}{
    \begin{tabular}{l c c c c}
    \toprule
    \textbf{Method} & \textbf{Condition} & \textbf{FGD}$\downarrow$ & \textbf{BC}$\uparrow$ & \textbf{Diversity}$\uparrow$ \\
    \midrule
    DisCo              & Audio        & 9.417 & 6.439 & 9.91 \\
    CaMN               & Audio, Text  & 6.644 & 6.769 & 10.86 \\
    TalkSHOW           & Audio        & 6.209 & 6.947 & 13.47 \\
    EMAGE              & Audio, Text  & 5.512 & 7.724 & 13.06 \\
    SynTalker          & Audio, Text  & 6.413 & 7.971 & 12.72 \\
    Language-of-Motion  & Audio        & 5.301 & 7.780 & \textbf{15.17} \\
    \midrule
    \textbf{SentiAvatar (Ours)} & Audio, Text & \textbf{4.941} & \textbf{8.078} & 10.56 \\
    \rowcolor{gray!10} \multicolumn{2}{l}{\textbf{Improvement (\%)}} & \textbf{+6.8} & \textbf{+1.3} & -- \\
    \bottomrule
    \end{tabular}}
\end{table}

\begin{table}[t]
    \centering
    \caption{User study results (1--5 Likert scale, higher is better).}
    \label{tab:user_study}
    \scalebox{0.88}{
    \begin{tabular}{l c c c c}
    \toprule
    \textbf{Method} & \textbf{Condition} & \textbf{Overall}$\uparrow$ & \textbf{Semantic}$\uparrow$ & \textbf{Prosody}$\uparrow$ \\
    \midrule
    HunYuan-Motion  & Text & 1.88 & 2.09 & 2.01 \\
    MoMask  &  Text     & 2.67 & 2.79 & 2.76 \\
    EMAGE    &   Audio    & 2.48 & 2.44 & 2.70 \\
    AT2M-GPT  & Audio, Text   & 2.33 & 2.37 & 2.45 \\\midrule
    \textbf{SentiAvatar (Ours)}& Audio, Text  & \textbf{2.99} & \textbf{2.97} & \textbf{3.16} \\
    \rowcolor{gray!10} \multicolumn{2}{l}{\textbf{Improvement (\%)}} & \textbf{+12.0} & \textbf{+6.5} & \textbf{+14.5} \\
    \bottomrule
    \end{tabular}}
\end{table}

\subsection{Main Results}
\label{sec:main_results}
\paragraph{Results on SuSuInterActs.}
Table~\ref{tab:susu_main} presents the quantitative comparison. SentiAvatar achieves an R@1 of 43.64\%, nearly 2$\times$ the best text-only baseline MoMask (34.55\%) and 8$\times$ EMAGE (5.00\%), demonstrating that our LLM planner effectively translates motion labels into semantically consistent keyframe plans.
Our FID (8.912) is 4$\times$ lower than MoMask (36.25) and 49$\times$ lower than EMAGE (441.6), indicating the generated distribution closely matches ground truth.
For audio--motion synchronization, SentiAvatar achieves the best ESD (0.456s) among all methods, outperforming EMAGE (0.606s) and text-only methods (T2M-GPT: 0.721s), confirming that the Infill Transformer captures fine-grained speech--motion correspondence.
HunYuan-Motion, despite being a large-scale model, scores poorly across all metrics (R@1 5.21\%, FID 352.56) as a zero-shot baseline not trained on motion-capture data with our skeleton topology.
Diversity (22.41) is comparable to real motion (22.61), indicating the pre-trained foundation model prevents mode collapse.

\paragraph{Results on BEATv2.}
Table~\ref{tab:beatv2_main} shows that SentiAvatar achieves the best FGD (4.941) and BC (8.078) on this benchmark, improving the prior best results (FGD 5.301, BC 7.971). This confirms that our architecture generalizes across languages and datasets. The lower diversity (10.56) is expected: our model generates audio-conditioned motion rather than sampling freely from the prior.

\paragraph{User Study.}
\label{sec:user_study}
We conduct a user study where 10 participants rate videos on a 1--5 Likert scale across three dimensions: semantic consistency, prosody synchronization, and overall quality. As shown in Table~\ref{tab:user_study}, SentiAvatar achieves the highest scores on all three dimensions. The prosody score (3.16) shows the largest margin over baselines, reflecting the benefit of our audio-conditioned Infill Transformer. MoMask achieves the second-best semantic score (2.79) but lags in prosody (2.76) due to its text-only design. EMAGE scores well on prosody (2.70) but poorly on semantics (2.44), confirming that audio-only methods cannot capture action intent.

\begin{figure*}[t]
    \centering
    \includegraphics[width=\linewidth]{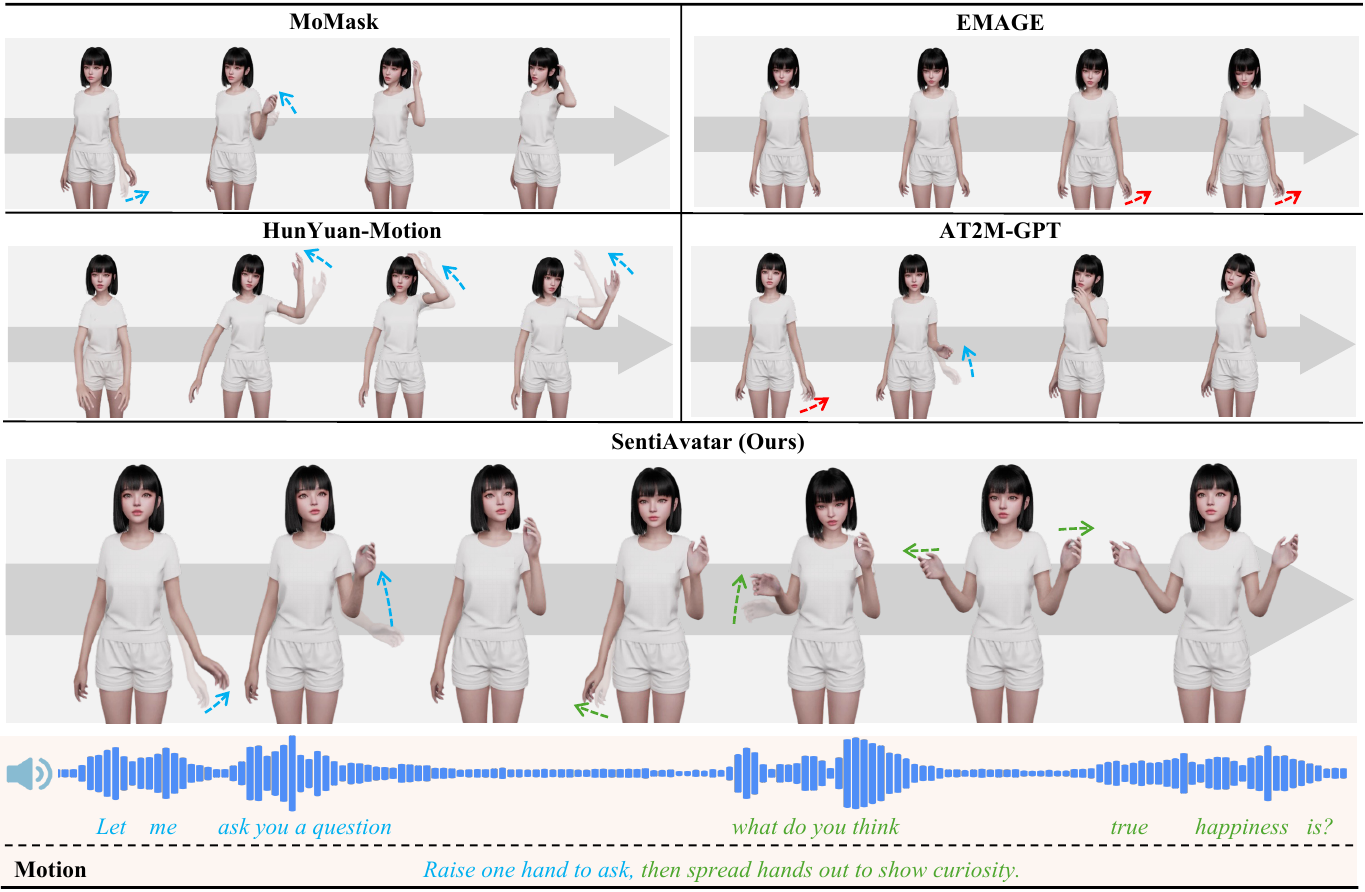}
    \caption{Qualitative comparison of generated motions across methods. Each row shows keyframe sequences for a given motion and speech. Texts and arrows of the same color indicate the same time step. The red arrow indicates an incorrect action.}
    \label{fig:qualitative}
\end{figure*}

\subsection{Qualitative Analysis}
\label{sec:qualitative}

Figure~\ref{fig:qualitative} compares keyframe sequences across methods. \textbf{SentiAvatar} produces motions that are both semantically correct (matching the intended action) and temporally aligned with the audio waveform, yielding the most natural results. \textbf{MoMask} partially captures action semantics from text labels but generates static-tempo motion with no audio correspondence, as it has no access to speech. \textbf{EMAGE} produces audio-synchronized movements---gestures track speech rhythm---but the motions are generic and ignore the semantic intent specified by the label. \textbf{AT2M-GPT}, despite receiving both audio and text, often misinterprets the action semantics compared to our hierarchical design, though its motion rhythm roughly follows the audio. \textbf{HunYuan-Motion} produces the lowest-quality output, with visible body distortions and unnatural poses; this is expected since it was not trained on high-quality motion-capture data, further highlighting the value of the SuSuInterActs dataset for character-specific interaction.

\begin{table}[t]
    \centering
    \caption{Architecture ablation on SuSuInterActs.}
    \label{tab:arch_ablation}
    \scalebox{0.92}{
    \begin{tabular}{l c c c c}
    \toprule
    \textbf{Variant} & \textbf{R@1}$\uparrow$ & \textbf{FID}$\downarrow$ & \textbf{ESD}$\downarrow$ & \textbf{Div.}$\uparrow$ \\
    \midrule
    w/o Pre-training          & \underline{42.56} & 8.988 & \underline{0.452} & \textbf{22.42} \\
    w/o Infill Transformer    & 27.52 & 18.491 & 0.503 & 22.36 \\
    w/o LLM Planner           & 28.06 & 27.567 & \textbf{0.421} & 22.33 \\
    Full pipeline              & \textbf{43.64} & \textbf{8.912} & 0.456 & \underline{22.41} \\
    \bottomrule
    \end{tabular}}
\end{table}

\begin{table}[t]
    \centering
    \caption{Ablation on audio conditioning.}
    \label{tab:audio_ablation}
    \scalebox{0.92}{
    \begin{tabular}{l c c c c}
    \toprule
    \textbf{Variant} & \textbf{R@1}$\uparrow$ & \textbf{FID}$\downarrow$ & \textbf{ESD}$\downarrow$ & \textbf{Div.}$\uparrow$ \\
    \midrule
    w/o all audio           & 41.72 & 9.996 & 0.523 & 22.37 \\
    w/o audio (Infill)      & 41.41 & 10.165 & \underline{0.497} & 22.31 \\
    w/o audio (LLM)         & \underline{43.48} & \underline{9.690} & 0.517 & \textbf{22.52} \\
    Full (both audio)       & \textbf{43.64} & \textbf{8.912} & \textbf{0.456} & \underline{22.41} \\
    \bottomrule
    \end{tabular}}
\end{table}

\begin{table*}[t]
    \centering
    \caption{Ablation on keyframe interval (step $t$).}
    \label{tab:step_ablation}
    \scalebox{1.0}{
    \begin{tabular}{c c c c c}
    \toprule
    \textbf{Step $t$} & \textbf{R@1}$\uparrow$ & \textbf{FID}$\downarrow$ & \textbf{ESD}$\downarrow$ & \textbf{Div.}$\uparrow$ \\
    \midrule
    Token-by-Token  & 27.52 & 18.491 & 0.503 & 22.36 \\\midrule
    $t{=}2$  & 34.21 & 11.773 & 0.472 & 22.16 \\
    $t{=}4$  & \textbf{43.64} & \textbf{8.912} & \underline{0.456} & \underline{22.41} \\
    $t{=}8$  & \underline{36.44} & \underline{9.205} & \textbf{0.439} & \textbf{22.63} \\
    \bottomrule
    \end{tabular}}
\end{table*}

\subsection{Ablation Study}
\label{sec:ablation}

\paragraph{Architecture ablation.}
Table~\ref{tab:arch_ablation} validates each component. Removing the LLM planner causes the steepest drop: R@1 falls from 43.64\% to 28.06\% and FID degrades from 8.912 to 27.567, confirming that sentence-level semantic planning is essential. Its ESD (0.421) is slightly better because the Infill Transformer directly conditions on every audio frame without the keyframe bottleneck, but this local alignment comes at the cost of global semantic coherence. Removing the Infill Transformer degrades all metrics (R@1 27.52\%, FID 18.491, ESD 0.503), as sparse keyframes alone produce jerky motion with poor temporal alignment. Pre-training contributes a modest but consistent gain (R@1 42.56\% $\rightarrow$ 43.64\%), indicating that the diverse 200K-sequence prior complements task-specific fine-tuning. The full pipeline achieves the best overall balance.

\paragraph{Audio conditioning.}
Table~\ref{tab:audio_ablation} disentangles the contribution of audio at each stage. Audio in the Infill Transformer is the primary driver of synchronization: removing it worsens ESD from 0.456s to 0.497s, and removing all audio further degrades ESD to 0.523s. Audio in the LLM mainly improves motion quality (FID 9.690 $\rightarrow$ 8.912) and rhythm planning (ESD 0.517 $\rightarrow$ 0.456), while text labels carry most semantic content (R@1 drops only 0.16\%). Full audio conditioning at both stages yields the best overall performance, confirming the synergy between coarse-grained audio planning and fine-grained audio alignment.

\paragraph{Keyframe interval.}
Table~\ref{tab:step_ablation} studies the step $t$ trade-off. Dense planning ($t{=}2$) overloads the LLM with long token sequences, degrading R@1 to 34.21\%. Sparse planning ($t{=}8$) achieves the best ESD (0.439s) and diversity (22.63) since the Infill Transformer has more freedom, but semantic alignment drops (R@1 36.44\%). The default $t{=}4$ strikes the best balance (R@1 43.64\%, FID 8.912). Token-by-token generation without the Infill Transformer performs worst overall, confirming the advantage of our plan-then-infill design.

\section{Conclusion}
\label{sec:conclusion}
In this paper, we presented \textbf{SentiAvatar}, a novel framework for building highly expressive, interactive 3D digital humans, demonstrated through our real-time character, SuSu. To address the scarcity of interactive conversational data and the challenges of motion-prosody synchronization, we introduced \textbf{SuSuInterActs}, a 37-hour multimodal dialogue corpus, and proposed a \textbf{plan-then-infill} architecture powered by a pre-trained Motion Foundation Model on over 200K sequences. This design effectively decouples high-level semantic planning from frame-level audio-driven interpolation, ensuring that the generated motions are both contextually accurate and rhythmically aligned with speech. Experiments demonstrate that SentiAvatar achieves state-of-the-art performance on both SuSuInterActs and BEATv2. With highly efficient generation (0.3 s for 6 s of output) and open-sourced resources, our work provides a robust foundation for future research in natural, real-time virtual human interactions.
\section*{Acknowledgment}
The authors would like to sincerely thank all collaborators for their valuable contributions to this work. In particular, special thanks to Shi Xueliang and Pan Xuanyue for leading the art design and data production efforts. The project also benefited greatly from the contributions of team members: Shi Xueliang, Yu Yongchang, Li Xing, and Liu Xueying in art design; Pan Xuanyue, Li Huixian, Yang Yijia, Zhang Wenxuan, and Wang Wei (UE) in data production. Their dedicated work and collaboration were essential to the successful completion of this research.

\bibliographystyle{unsrt}
\bibliography{sample-base}

\appendix
\appendix
\newpage

\section{Character Profile}
\label{app:character}

SuSu is designed as a cohabiting companion character whose personality balances warmth with playful reserve.
Table~\ref{tab:character_basic} and Table~\ref{tab:character_personality} detail her basic attributes and behavioral design, respectively.
These specifications guide both script generation and actor performance, ensuring consistent and diverse behavioral data.

\begin{table}[h]
\centering
\small
\begin{tabular}{l p{5.8cm}}
\toprule
\textbf{Attribute} & \textbf{Description} \\
\midrule
Name        & SuSu (user's cohabiting roommate) \\
Age         & 22 \\
Height      & 165\,cm \\
Education   & B.A.\ in Chinese Language \& Literature; strong writing skills \\
Build       & Slender; long limbs, narrow shoulders; movements carry a dance-like fluidity \\
Appearance  & Oval face, almond eyes with naturally curled lashes, straight nose bridge, defined jawline. Fair, cool-toned skin. Expressions shift with a slight ``delay,'' yielding a soft, natural quality. Gaze conveys layered emotion---calm on the surface, with ripples beneath. \\
\bottomrule
\end{tabular}
\caption{SuSu: basic attributes.}
\label{tab:character_basic}
\end{table}

\begin{table}
\centering
\small
\begin{tabular}{l p{5.8cm}}
\toprule
\textbf{Aspect} & \textbf{Description} \\
\midrule
Personality 
  & Quiet and slow to warm up, yet takes initiative once comfortable. Gentle with a playful edge; sweet-tongued without being cloying. Stubborn about small things; occasionally over-serious. Witty and tactful; feigns composure. \\
\addlinespace
Language style 
  & Concise and colloquial. Serious when discussing real matters; slips in dry humor otherwise. Maintains a push-pull dynamic---neither too close nor too distant---conveying affection through indirection, light teasing, and implicit hints rather than explicit statements. \\
\addlinespace
Social habits 
  & Introverted homebody; reluctant to go out, yet naturally warm and approachable in conversation. \\
\addlinespace
Core trait 
  & An inner tension between sensitivity and desire for connection. Uses politeness and emotional distance as self-protection, yet reveals wit, playfulness, and longing for closeness in text and virtual interactions. Interaction pattern: tentative approach then retreat, conveying affection through push-and-pull. \\
\addlinespace
Motion logic 
  & Movements mirror the push-pull core: approach then withdraw, touch then retract. \emph{Happy}: gentle smile building to laughter, usually a subtle grin. \emph{Angry}: face goes cold for self-protection, then turns back with a surprise softening once the other party apologizes. \emph{Sad}: speaks something genuinely moving, then laughs it off as a joke. Every gesture is designed to convey feminine charm and emotional nuance. \\
\addlinespace
Speech rules 
  & Never answers a question head-on; uses metaphor, exaggeration, wordplay, rhetorical questions, or unexpected angles. Each utterance is 10--30 Chinese characters, colloquial and easy to understand. Tone is light and keeps conversations from becoming heavy. \\
\bottomrule
\end{tabular}
\caption{SuSu: personality and behavioral design. These specifications guide script generation and actor performance to produce diverse yet character-consistent data.}
\label{tab:character_personality}
\end{table}

\section{LLM Continuation Mode}
\label{app:continuation}

To support continuous generation across utterances, the LLM operates in a \emph{continuation mode}. The model receives a short prefix of audio--motion token pairs from the end of the previous utterance, followed by the motion label and new audio tokens for the current utterance:
\begin{align}
\text{Input:} &\quad [a_{1}][a_{1+t}][\mathbf{r}_{1}][\mathbf{r}_{1+t}] \oplus \mathbf{T} \oplus [a_{1+2t}][a_{1+3t}] \cdots, \\
\text{Output:} &\quad [\text{len}][\mathbf{r}_{1+2t}][\mathbf{r}_{1+3t}] \cdots,
\end{align}
where the prefix contains the last two keyframe audio--motion pairs from the previous turn, and generation starts from the next keyframe position after the context boundary.

During training, since paired consecutive utterances are not always available, we simulate the context by splitting a single utterance: the first few keyframes serve as pseudo-history, while the remaining form the generation target. At inference, the actual last keyframes from the previous utterance are used as the prefix, enabling smooth cross-utterance transitions. The Infill Transformer's sliding window naturally handles the boundary: the last keyframe of the previous utterance becomes the known start frame of the first interpolation window for the new utterance.

\section{Hunyuan Motion Model Distillation}
\label{app:hunyuan}

To enrich the motion prior with diverse action semantics, we distill 15K sequences from the Tencent Hunyuan Motion Model through a four-step prompt engineering pipeline.

\paragraph{Step 1: Atomic action mining.}
We collect common English verbs from dictionaries and test whether the Hunyuan model generates semantically correct motion from a single-word prompt. A verb is accepted if its output is consistent with the motion produced by Hunyuan's built-in rewrite module. Verbs producing imperceptible movements or purely facial expressions are discarded.

\paragraph{Step 2: Synonym expansion.}
Each accepted atomic action is expanded into synonymous phrases via an LLM. All phrases are verified to be recognizable by the Hunyuan model.

\paragraph{Step 3: Composite motion generation.}
Atomic actions and phrases are composed into longer descriptions (up to 4 actions) using temporal connective templates (start / transition / concurrency / ending), generated by GPT-4o-mini with physical plausibility constraints (e.g., limb occupancy logic). This yields $\sim$10K composite motion descriptions.

\paragraph{Step 4: Specialized categories.}
We supplement the dataset with Olympic sports movements (e.g., diving, martial arts), creature imitation actions, and path-modified base actions (e.g., walking in figure-eight patterns). After filtering, $\sim$5K additional samples are retained.

All distilled sequences are 5\,s in length. Combined with the three open-source datasets, the total pre-training corpus comprises 200K+ sequences and 676 hours (Table~\ref{tab:pretrain_data}).

\section{Event Sync Distance (ESD) Metric}
\label{app:esd}

We introduce \emph{Event Sync Distance} (ESD) as a bidirectional, event-level metric for evaluating audio--motion temporal synchronization. Unlike traditional beat alignment scores that compute only unidirectional distances (e.g., from motion peaks to the nearest audio beat), ESD considers both directions and is thus sensitive to both missed synchronization (poor recall) and spurious motion events (poor precision).

\subsection{Event Extraction}

\paragraph{Audio events.}
We extract audio onset events using librosa's beat tracking pipeline:
(1)~Convert the audio waveform to a Mel spectrogram;
(2)~Compute the onset strength envelope, which captures frames where spectral energy increases sharply---corresponding to rhythmic accents, phoneme onsets, or prosodic stress in speech;
(3)~Estimate the global tempo via autocorrelation;
(4)~Use dynamic programming to select a set of onset positions that best fit the estimated tempo.
The resulting frame indices are converted to seconds, yielding audio event times $A = \{a_1, a_2, \dots, a_n\}$.

\paragraph{Motion events.}
We detect kinematic ``burst'' points---moments of peak limb velocity---via the following steps:
(1)~Compute inter-frame displacement: $\Delta \mathbf{x}_t = \mathbf{x}_{t+1} - \mathbf{x}_t$;
(2)~Compute velocity magnitude: $v_t = \|\Delta \mathbf{x}_t\|_2$;
(3)~Apply a dynamic amplitude threshold to filter out micro-jitter caused by generation instability:
\begin{equation}
    \text{Threshold} = \mu_v + 0.2 \cdot \sigma_v,
\end{equation}
where $\mu_v$ and $\sigma_v$ are the global mean and standard deviation of the velocity sequence;
(4)~Detect local maxima (peaks) in the thresholded velocity curve;
(5)~Convert peak frame indices to seconds using the motion frame rate.
This yields motion event times $M = \{m_1, m_2, \dots, m_k\}$.

\subsection{Metric Computation}

Given audio events $A$ and motion events $M$, we first construct the pairwise absolute time-difference matrix:
\begin{equation}
    D_{i,j} = |a_i - m_j|, \quad i \in \{1, \dots, n\},\; j \in \{1, \dots, k\}.
\end{equation}

\paragraph{Audio-to-motion distance (recall).}
For each audio event, find the nearest motion event:
\begin{equation}
    d_{A \to M} = \frac{1}{n} \sum_{i=1}^{n} \min_{j \in \{1, \dots, k\}} D_{i,j}.
\end{equation}
This measures how well audio events are ``covered'' by nearby motion events. A high $d_{A \to M}$ indicates that many audio onsets lack corresponding motion responses.

\paragraph{Motion-to-audio distance (precision).}
For each motion event, find the nearest audio event:
\begin{equation}
    d_{M \to A} = \frac{1}{k} \sum_{j=1}^{k} \min_{i \in \{1, \dots, n\}} D_{i,j}.
\end{equation}
This measures whether each motion event is temporally justified by a nearby audio event. A high $d_{M \to A}$ indicates that the model produces many spurious or randomly-timed motion bursts.

\paragraph{ESD score.}
The final score is the average of both directions:
\begin{equation}
    \text{ESD} = \frac{d_{A \to M} + d_{M \to A}}{2}.
\end{equation}
ESD is measured in seconds; lower values indicate better audio--motion synchronization. When either $A$ or $M$ is empty (no events detected), a penalty of 2.0s is assigned.

\subsection{Design Rationale}

The bidirectional formulation addresses two failure modes that unidirectional metrics miss:
\begin{itemize}
    \item \textbf{Over-active motion} (e.g., jittering or random flailing): produces many motion events, so $d_{A \to M}$ may be artificially low (every audio event happens to be near \emph{some} motion peak), but $d_{M \to A}$ will be high because most motion peaks have no corresponding audio justification.
    \item \textbf{Under-active motion} (e.g., near-static poses): $d_{M \to A}$ may be low (the few motion events that exist are near audio events), but $d_{A \to M}$ will be high because most audio events are not covered.
\end{itemize}
By averaging both directions, ESD penalizes both failure modes equally, providing a more robust synchronization assessment than unidirectional metrics.

\end{document}